\documentclass[11pt]{article}
\usepackage{amsmath,amsfonts,amsthm,amssymb,color}
\usepackage{graphicx,amssymb,amstext,amsmath,amsthm}
\usepackage{fancybox}
\usepackage{multirow}

\usepackage{times}
\usepackage{helvet}
\usepackage{courier}
\usepackage{url}  
\usepackage{graphicx,amssymb,amstext,amsmath,amsthm}
\usepackage{subfig}
\usepackage[algoruled,linesnumbered]{algorithm2e}
\usepackage{wrapfig}
\usepackage{color}

\usepackage{ifthen}

\newboolean{@full}
\setboolean{@full}{true}  

\newcommand{\iffull}{\ifthenelse{\boolean{@full}}}

\newtheorem{theorem}{Theorem}[section]

\newtheorem{lemma}[theorem]{Lemma}

\theoremstyle{definition}

\newenvironment{fminipage}%
  {\begin{Sbox}\begin{minipage}}%
  {\end{minipage}\end{Sbox}\fbox{\TheSbox}}

\iffull{
	\textheight 8.5in
	\topmargin -0.2in
	\oddsidemargin 0.20in
	\textwidth 6.3in
	
	\newenvironment{algbox}[0]{\vskip 0.2in
	\noindent 
	\begin{fminipage}{6.3in}
	}{
	\end{fminipage}
	\vskip 0.2in
	}
}{
	\usepackage[margin=1in]{geometry}
	\usepackage{paralist}

	\setlength{\abovecaptionskip}{5pt}
	\setlength{\belowcaptionskip}{-15pt}
	
	\expandafter\def\expandafter\normalsize\expandafter{%
	    \normalsize
	    \setlength\abovedisplayskip{5pt}
	    \setlength\belowdisplayskip{5pt}
	    \setlength\abovedisplayshortskip{5pt}
	    \setlength\belowdisplayshortskip{5pt}
	}

	\usepackage{comment}
	\excludecomment{proof}
	
	\usepackage{titlesec}
	\titlespacing*{\section}{0pt}{0.2\baselineskip}{0.2\baselineskip}
	\titlespacing*{\subsection}{0pt}{0.2\baselineskip}{0.2\baselineskip}
	\titlespacing*{\theorem}{0pt}{0\baselineskip}{0\baselineskip}
	\titlespacing*{\lemma}{0pt}{0\baselineskip}{0\baselineskip}
}

\newcommand{\todo}[1]{}
\newcommand{\yintat}[1]{}
\newcommand{\dan}[1]{}
\newcommand{\richard}[1]{}

\usepackage{thm-restate}

\begin{document}

\title{Multi-Entity Dependence Learning with Rich Context via \\
Conditional Variational Auto-encoder}

\author{
		Luming Tang$^*$ \\
		tlm14@mails.tsinghua.edu.cn\\
		Tsinghua University
		\and
		Yexiang Xue\thanks{indicates equal contribution}\\
		yexiang@cs.cornell.edu\\
		Cornell University
		\and
		Di Chen \\
		dc874@cornell.edu\\
		Cornell University		
		\and
		Carla P. Gomes\\
		gomes@cs.cornell.edu\\
		Cornell University
	}
	\maketitle

\begin{abstract}
Multi-Entity Dependence Learning ($\mathtt{MEDL}$) explores
conditional correlations among multiple entities. The availability of
rich contextual information requires a nimble learning scheme that
tightly integrates with deep neural networks and has the ability to
capture correlation structures among exponentially many outcomes. We propose $\mathtt{MEDL\_CVAE}$, which encodes a conditional multivariate distribution as a generating process. As a result, the variational lower bound of the joint likelihood can be optimized via a conditional variational auto-encoder and trained end-to-end on GPUs. Our $\mathtt{MEDL\_CVAE}$ was motivated by two real-world applications in computational sustainability: one studies the spatial correlation among multiple bird species using the \emph{eBird} data and the other models multi-dimensional landscape composition and human footprint in the Amazon rainforest with satellite images. We show that $\mathtt{MEDL\_CVAE}$ captures rich dependency structures, scales better than previous methods, and further improves on the joint likelihood taking advantage of very large datasets that are beyond the capacity of previous methods.
\end{abstract}

\section{Introduction}

Learning the dependencies among multiple entities is an important
problem with many real-world applications.
For example, in the sustainability domain, the spatial 
distribution of one species depends on other species due to their interactions in the form of mutualism, commensalism, competition and predation \cite{MacKenzie2004cooccurrence}.
In natural language processing, the
topics of an article are often correlated \cite{nam2014large}.
In computer vision, an image may have multiple correlated tags \cite{wang2016cnn}.

The key challenge behind dependency learning 
is to capture correlation structures embedded among exponentially many
outcomes.
One classic approach is the Conditional Random Fields (CRF) \cite{Lafferty2001CRF}.
However, to handle the intractable partition function resulting from
multi-entity interactions, CRFs have to incorporate approximate
inference techniques such as contrastive divergence \cite{hinton2002training}.
In a related applicational domain called multi-label classification,
the classifier
chains (CC) approach \cite{Read09ClassifierChain} decomposes the joint likelihood into a product
of conditionals and reduces a multi-label classification problem to a
series of binary prediction problems.
However, as pointed out by \cite{Dembczynski2010PCC}, finding the joint mode of CC is also
intractable, and to date only approximate search methods are available \cite{dembczynski2012label}.

\begin{figure}[tb]
\centering
\includegraphics[width=0.6\linewidth]{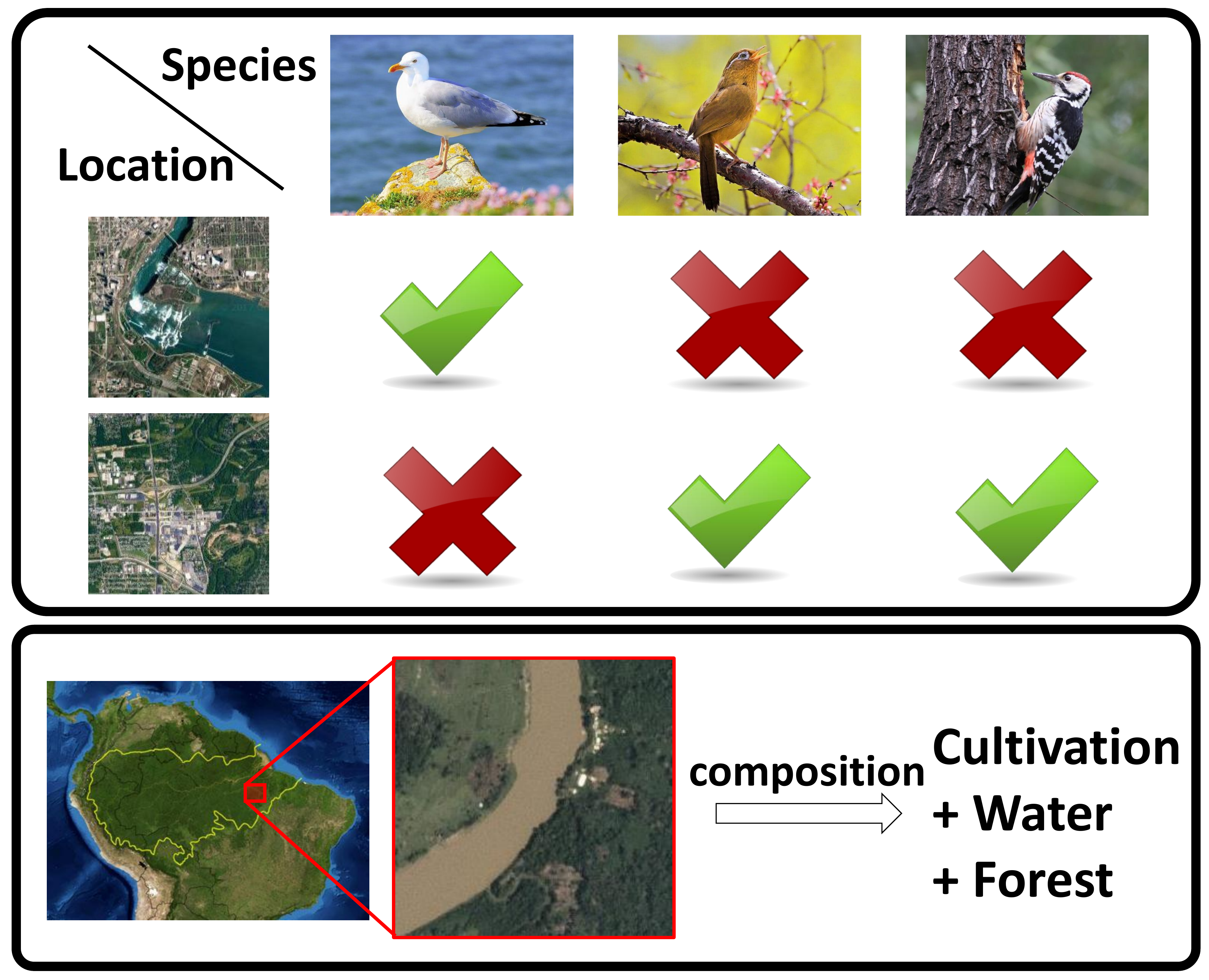}
\caption{Two computational sustainability related applications for
  $\mathtt{MEDL\_CVAE}$. The first application is to
  study the interactions among bird species in the crowdsourced
  \textit{eBird} dataset and environmental covariates including those
  from satellite images. The second application is to tag
  satellite images with a few potentially overlapping landscape
  categories and track human footprint in the Amazon rainforest.
}
\label{fig:intro}
\end{figure}

The availability of rich contextual information such as millions of
high-resolution satellite images as well as recent developments 
in deep learning create both opportunities and challenges for
multi-entity dependence learning. In terms of opportunities, 
rich contextual information creates 
the possibility of improving predictive performance,
especially when it is combined with highly flexible deep neural
networks. 

The challenge, however, is to design a
nimble scheme that can both tightly integrate with
deep neural networks and capture correlation
structures among exponentially many outcomes.
Deep neural nets are commonly used to extract features from contextual information sources, and can effectively use highly parallel infrastructure such as GPUs. However, classical approaches for structured output, such as sampling, approximate inference and search methods, typically cannot be easily parallelized.

Our contribution is {\bf an end-to-end approach to multi-entity
dependence learning based on
a conditional variational auto-encoder, which handles high
dimensional space effectively, and can be tightly integrated with
deep neural nets to take advantages of rich contextual
information}. Specifically, (i) we propose a novel 
\emph{generating process} to encode the conditional 
multivariate distribution in multi-entity dependence learning, in which we bring in a set of hidden
variables to capture the randomness in the joint distribution.  (ii)
The novel conditional generating process allows us to work with
imperfect data, 
capturing noisy and potentially \emph{multi-modal responses}.  (iii) The generating process 
also allows us to encode the entire problem via \emph{a conditional
variational auto-encoder}, {tightly integrated with deep neural nets
and implemented end-to-end on GPUs}.
Using this approach, we are able to leverage rich contextual
information to 
enhance the performance of $\mathtt{MEDL}$ that is
beyond the capacity of previous methods.

We apply our \underline{M}ulti-\underline{E}ntity \underline{D}ependence \underline{L}earning via \underline{C}onditional \underline{V}ariational \underline{A}uto-\underline{e}ncoder ($\mathtt{MEDL\_CVAE}$) approach to {\bf two sustainability related
  real-world applications} \cite{gomes2009computational}. In the first
application, we study the interaction among multiple bird species with
crowdsourced \textit{eBird} data and 
environmental covariates including those from satellite images.
As an important sustainable development indicator,  
studying how species distribution changes over time helps us understand the effects
of climate change and conservation strategies. 
In our second application, we use high-resolution satellite imagery to
study multi-dimensional landscape composition and track human footprint in the
Amazon rainforest. 
See Figure \ref{fig:intro} for an overview of the two problems. 
Both applications study the correlations of multiple entities
and use satellite images as rich context information.  
We are able to show that our $\mathtt{MEDL\_CVAE}$ (i) {\bf captures rich correlation structures among entities}, therefore {\bf outperforming approaches that assume independence among entities} given contextual information; (ii) {\bf trains in an order of magnitude less time than previous methods} because the full pipeline implemented on GPUs; (iii) achieves a better joint likelihood by incorporating deep neural nets to {\bf take advantage of rich context information, namely satellite images}, which are beyond the capacity of previous methods.

\section{Preliminaries}

We consider modeling the dependencies 
among multiple entities on problems with rich contextual information. 
Our dataset consists of tuples $D=\{(x_i, y_i)| i=1,\ldots,N\}$, in which $x_i =
(x_{i,1},\ldots, x_{i,k}) \in \mathcal{R}^k$ is a high-dimensional
contextual feature vector, and $y_i = (y_{i,1},\ldots,y_{i,l})\in \{0,1\}^l$ is a sequence of $l$ indicator variables, 
in which $y_{i,j}$ represents whether the $j$-th entity is observed in
an environment characterized by covariates $x_i$. 
The problem is to learn a conditional joint distribution $Pr(y | x)$ which maximizes the conditional joint log likelihood over $N$ data points: 
$$\sum\limits_{i=1}^{N} \log Pr(y_i | x_i).$$

Multi-entity dependence learning is a general problem with many
applications. For example, in our species distribution
application where we would like to model the relationships of multiple
bird species, $x_i$ is the vector of environmental covariates of the
observational site, which includes a remote sensing picture, the
national landscape classification dataset (NLCD) values \cite{homer2015completion} etc.
$y_i = (y_{i,1},\ldots,y_{i,l})$ is a sequence of binary indicator
variables, where $y_{i,j}$ indicates whether species $j$ is
detected in the observational session of the $i$-th data point. In our application to analyze landscape
composition, $x_i$ is the feature vector made up
with the satellite image of the given site, and $y_i$ includes multiple
indicator variables, such as atmospheric conditions (clear or cloudy)
and land cover phenomena (agriculture or forest) of the site.

\begin{figure}[tb]
\centering
\includegraphics[width=0.6\linewidth]{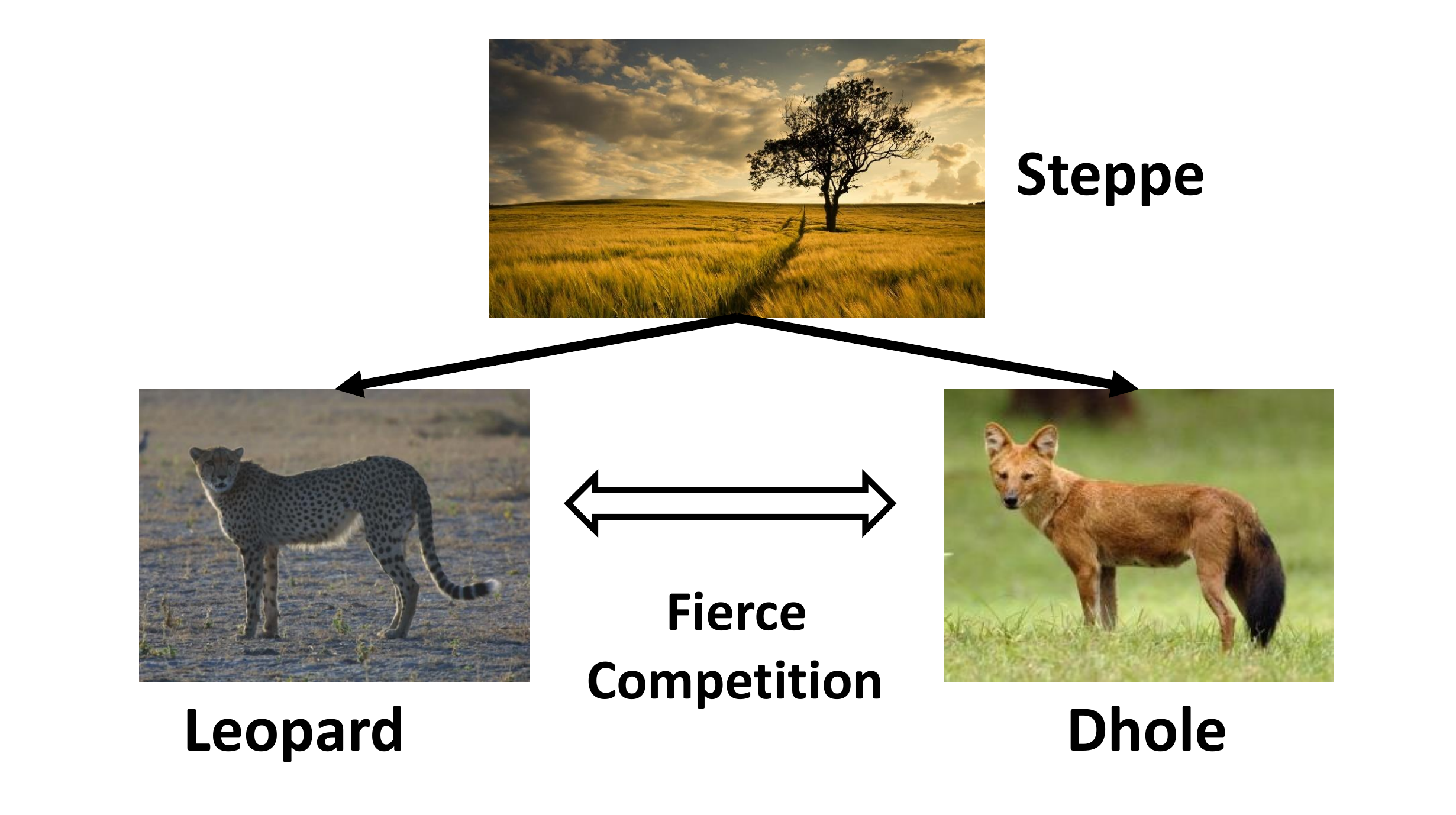}
\caption{Leopards and dholes both live on steppes. Therefore, the probability that each animal occupies a steppe is high. However, due to the competition between the two species, the probability of their co-existence is low.}
\label{fig:panther}
\end{figure}

Our problem is to capture rich correlations between entities.
For example, in our species distribution modeling application, the
distribution of multiple species are often correlated, due to their
interactions such as cooperation and competition for shared resources.
As a result, we often cannot assume that the probability of each entity's existence are mutually
independent given the feature vector, i.e.,
\begin{equation}
  Pr(y_i|x_i) \neq \prod_{j=1}^l Pr(y_{i,j}|x_i).
  \label{eq:ind}
\end{equation}
See Figure \ref{fig:panther} for a specific instance. 
As a baseline, we call the model which takes the
assumption in the righthand side of Equation~(\ref{eq:ind}) an independent probabilistic model.

\section{Our Approach}

We propose $\mathtt{MEDL\_CVAE}$ to address two
challenges in multi-entity dependence learning. 

The first challenge is the noisy and potentially multi-modal responses. 
For example, consider our species distribution modeling application. 
One bird watcher can make slightly different observations
during two consecutive visits to the same forest location. He may be able to detect a song bird 
during one visit but not the other if, for example, the bird does not sing both times.
This suggests that, even under the very best effort of bird watchers, there is still noise inherently associated with the observations.
Another, perhaps more complicated phenomenon is the multi-modal
response, which results from intertwined ecological processes such as
mutualism and commensalism.
Consider, territorial species such as the Red-winged and Yellow-headed Blackbirds, both of which live in open marshes in Northwestern United States. However, the Yellowheads tend to chase the Redwings out of their territories. 
As a result, a bird watcher would see either Red-winged or Yellow-headed Blackbirds at an open marsh, but seldom both of them. This suggests that, conditioned on an open marsh environment, there are  two possible modes in the response, seeing the Red-winged but not the Yellow-headed, or seeing the Yellow-headed but not the Red-winged.

The second challenge comes from the incorporation of rich contextual information such as remote sensing imagery, wich provides detailed feature description of the underlying environment, especially in conjunction with the flexibility of deep neural networks.
Nevertheless, previous multi-entity models, such as in \cite{chen2016deep,Guo2011MultiCDN,Sutton2012CRF}, rely on sampling approaches to estimate the partition function during training.  
%
It is difficult to incorporate such sampling process into the back-propagation of the deep neural networks.
This limitation poses serious challenges to taking advantage of the rich contextual information. 

To address the aforementioned two challenges, we propose a conditional generating model, which makes use of hidden variables to represent the noisy and multi-modal responses.
$\mathtt{MEDL\_CVAE}$ incorporates this model into an end-to-end training pipeline using a conditional variational autoencoder, which optimizes for a variational lower bound of the conditional likelihood function.

\begin{figure}[t]
    \begin{minipage}[t]{0.45\linewidth}
        \includegraphics[width=1.0\linewidth]{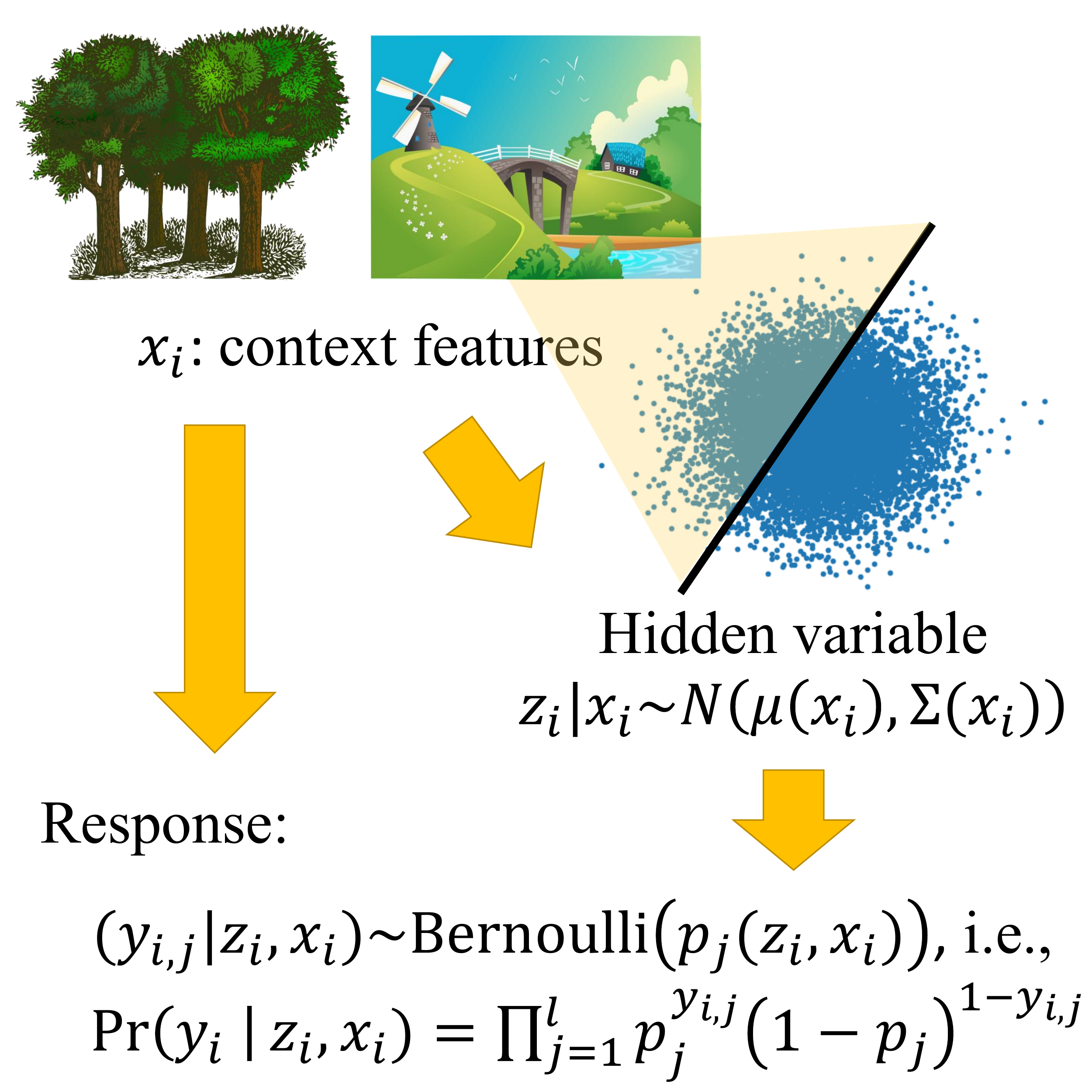}
        \captionsetup{labelformat=empty}
        \caption{(a) conditional generating process}
    \end{minipage}%
    \begin{minipage}[t]{0.54\linewidth}
        \centering
        \includegraphics[width=1.0\linewidth]{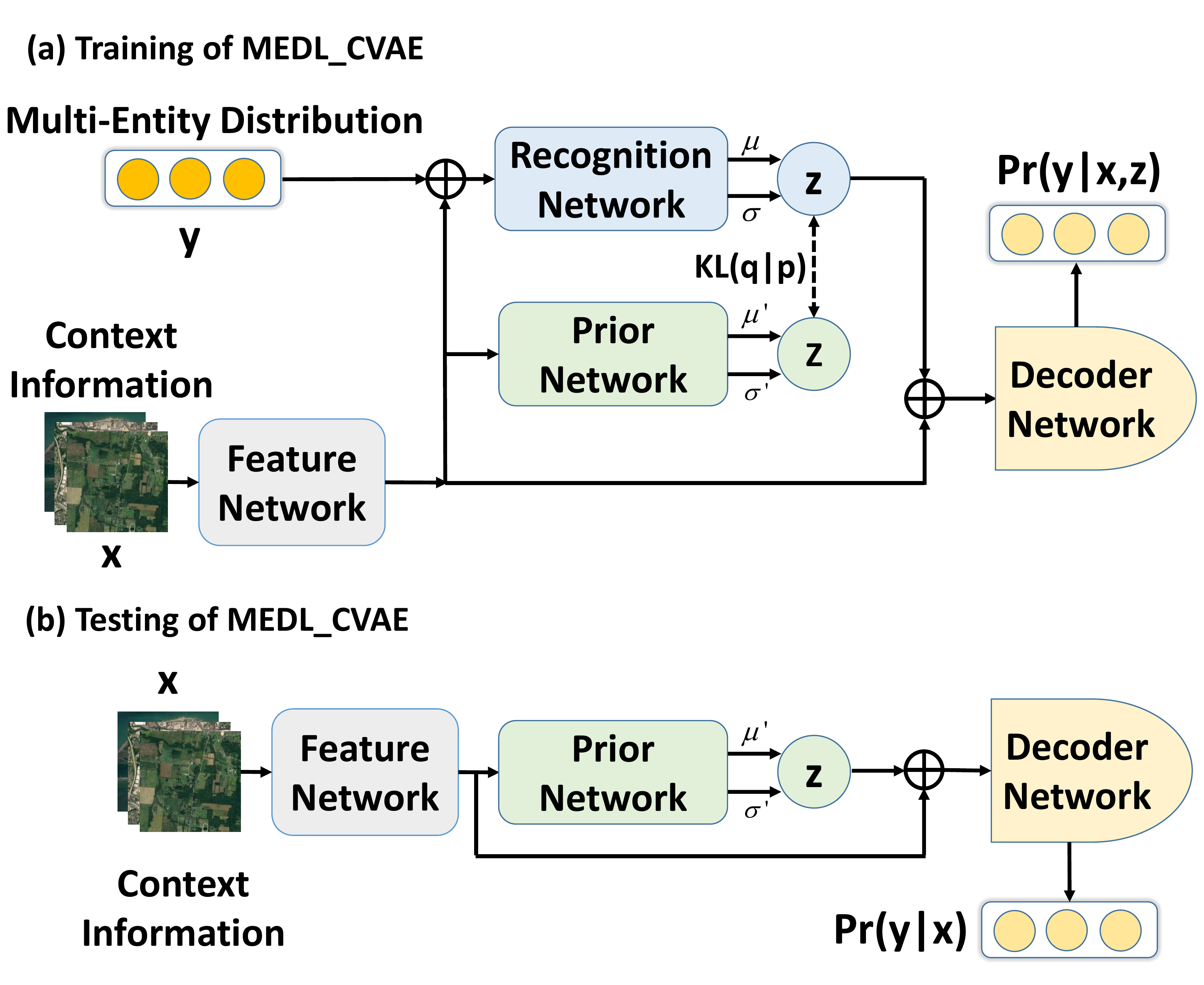}
        \captionsetup{labelformat=empty}
        \caption{(b) neural network architecture}
    \end{minipage}
    \caption{(a) Our proposed conditional generating process. Given contextual features $x_i$ such as satellite images, we use hidden variables $z_i$ conditionally generated based on $x_i$ to capture noisy and multi-modal response. The response $y_i$ depends on both contextual information $x_i$ and hidden variables $z_i$. See the main text for details. (b) Overview of the neural network architecture of $\mathtt{MEDL\_CVAE}$ for both training and testing stages. $\oplus$ denotes a concatenation operator.}
    \label{fig:method}
\end{figure}

\subsubsection{Conditional Generating Process}

Unlike classic approaches such as probit models \cite{chib1998analysis}, which have a single mode, we use a conditional generating process, which models noisy and multi-modal responses using additional hidden variables. The generating process is depicted in Figure.~\ref{fig:method}. 

In the generating process, we are given contextual features $x_i$, which for example, contain a satellite image. Then we assume a set of hidden variables $z_i$, which are generated based on a normal distribution conditioned on the values of $x_i$. The binary response variables $y_{i,j}$ are drawn from a Bernoulli distribution, whose parameters depend on both the contextual features $x_i$ and hidden variables $z_i$. The complete generating process becomes, 
\begin{align}
  x_i &: \mbox{contextual features},\\
  z_i|x_i &\sim N\left(\mu_d(x_i), \Sigma_d(x_i)\right),\\
  y_{i,j}|z_i,x_i &\sim \mbox{Bernoulli}\left(p_j(z_i,x_i)\right).
\end{align}
Here, $\mu_d(x_i)$, $\Sigma_d(x_i)$ and $p_j(z_i,x_i)$ are general functions depending on $x_i$ and $z_i$, which are modeled as deep neural networks in our application and learned from data. We denote the parameters in these neural networks as $\theta_d$. The machine learning problem is to find the best parameters that maximize the conditional likelihood $\prod_{i} Pr(y_i|x_i)$. 

This generating process is able to capture noisy and potentially multi-modal distributions. Consider the Red-winged and the Yellow-headed Blackbird example. We use $y_{i,1}$ to denote the occurrence of Red-winged Blackbird and $y_{i,2}$ to denote the occurrence of Yellow-headed Blackbird. Conditioned on the same environmental context $x_i$ of an open marsh, the output $(y_{i,1}=0, y_{i,2}=1)$ and $(y_{i,1}=1, y_{i,2}=0)$ should both have high probabilities. Therefore, there are two modes in the probability distribution.
Notice that it is very difficult to describe this case in any probabilistic model that assumes a single mode.

Our generating process provides the flexibility to capture multi-modal distributions of this type. 
The high-level idea is similar to mixture models, where we use hidden variables $z_i$ to denote which mode the actual probabilistic outcome is in. 
For example, we can have $z_i|x_i\sim N(0,I)$ and two functions $p_1(z)$ $p_2(z)$, where half of the $z_i$ values are mapped to $(p_1=0,p_2=1)$ and the other half to $(p_1=1,p_2=0)$. Figure~\ref{fig:method} provides an example, where the $z_i$ values in the region with a yellow background are mapped to one value, and the remaining values are mapped to the other value. In this way, the model will have high probabilities to produce both outcomes  $(y_{i,1}=0, y_{i,2}=1)$ and $(y_{i,1}=1, y_{i,2}=0)$. 
\subsubsection{Conditional Variational Autoencoder}

Our training algorithm is to maximize the conditional likelihood $Pr(y_i|x_i)$. Nevertheless, a direct method would result in the following optimization problem: 
\begin{small}
\begin{equation}
  \max_{\theta_d} \sum_i \log Pr(y_i|x_i) = \sum_i \log \int Pr(y_i|x_i,z_i) Pr(z_i|x_i) \mbox{d} z_i\nonumber
\end{equation}
\end{small}
which is intractable because of a hard integral inside the logarithmic function. 
Instead, we turn to maximizing variational lower bound of the conditional log likelihood. To do this, we use a variational function family $Q(z_i|x_i, y_i)$ to approximate the posterior: $Pr(z_i | x_i, y_i)$. 
In practice, $Q(z_i|x_i,y_i)$ is modeled using a conditional normal distribution:
\begin{equation}
Q(z_i|x_i,y_i) = N(\mu_e(x_i, y_i), \Sigma_e(x_i, y_i)).
\end{equation}
Here, $\mu_e(x_i, y_i)$ and $\Sigma_e(x_i, y_i)$ are general functions, and are modeled using deep neural networks whose parameters are denoted as $\theta_e$. We assume $\Sigma_e$ is a diagonal matrix in our formulation. 
Following similar ideas in \cite{kingma2013auto,kingma2014semi}, we can prove the following variational equality:
\begin{small}
\begin{align}
  &\log Pr(y_i | x_i) - D\left[Q(z_i|x_i,y_i) || Pr(z_i | x_i, y_i)\right]\label{eq:vari}\\
=&\mathtt{E}_{z_i\sim Q(z_i | x_i, y_i)}\left[\log Pr(y_i| z_i, x_i)\right] - 
   D\left[Q(z_i|x_i, y_i) || Pr(z_i | x_i)\right] \nonumber   
\end{align}
\end{small}
On the left-hand side, the first term is the conditional likelihood, which is our objective function. The second term is the Kullback-Leibler (KL) divergence, which measures how close the variational approximation $Q(z_i|x_i,y_i)$ is to the true posterior likelihood $Pr(z_i|x_i, y_i)$. Because $Q$ is modeled using a neural network, which captures a rich family of functions, we assume that $Q(z_i|x_i, y_i)$ approximates $Pr(z_i|x_i, y_i)$ well, and therefore the second KL term is almost zero. And because the KL divergence is always non-negative, the right-hand side of Equation~\ref{eq:vari} is a tight lower bound of the conditional likelihood, which is known as the variational lower bound.
We therefore directly maximize this value and the training problem becomes:
\begin{align}
\max_{\theta_d, \theta_e} ~~~~&\sum_i \mathtt{E}_{z_i\sim Q(z_i | x_i, y_i)}\left[\log Pr(y_i| z_i, x_i)\right] - \nonumber\\
   &D\left[Q(z_i|x_i, y_i) || Pr(z_i | x_i)\right].\label{eq:obj}
\end{align}

The first term of the objective function in Equation~\ref{eq:obj} can be directly formalized as two neural networks concatenated together -- one encoder network and the other decoder network, following the reparameterization trick,
which is used to backpropogate the gradient inside neural nets. 
At a high level, suppose $r\sim N(0, I)$ are samples from the standard Gaussian distribution, then $z_i\sim Q(z_i | x_i, y_i)$ can be generated from a ``recognition network'', which is part of the ``encoder network'': $z_i \leftarrow \mu_e(x_i, y_i) + \Sigma_e(x_i, y_i) r$.
The ``decoder network'' takes the input of $z_i$ from the encoder network and feeds it to the neural network representing the function $Pr(y_i|z_i, x_i) = \prod_{j=1}^l \left(p_j(z_i,x_i)\right)^{y_{i,j}} \left(1-p_j(z_i,x_i)\right)^{1-y_{i,j}}$ together with $x_i$. 
The second KL divergence term can be calculated in a close form. The entire neural network structure is shown as Figure \ref{fig:method}. We refer to $Pr(z|x)$ as the \textit{prior network}, $Q(z|x,y)$ as the \textit{recognition network} and $Pr(y|x,z)$ as the \textit{decoder network}. These three networks are all multi-layer fully connected neural networks. The fourth \textit{feature network}, composed of multi-layer convolutional or fully connected network, extracts high-level features from the contextual source.
All four neural networks are trained simultaneously using stochastic gradient descent. 


\section{Related Work}

Multi-entity dependence learning was studied extensively
for prediction problems under the names of multi-label classification
and structured prediction. Our applications, on the other hand, focus more on probabilistic modeling rather than classification. 
Along this line of research, early methods include k-nearest neighbors \cite{ZhouZhiHua2005KNearestMultiLabel} and dimension reduction \cite{ZhouZhihua2010MultiLabelDimensionReduct,Li2016MultiLabelBernM}.


\textbf{Classifier Chains} (CC) First proposed by \cite{Read09ClassifierChain}, the CC approach  decomposes the joint distribution into the product of a series of conditional probabilities. Therefore the multi-label classification problem is reduced to $l$ binary classification problems. As noted by \cite{Dembczynski2010PCC}, CC takes a greedy approach to find the joint mode and the result can be arbitrarily far from the true mode. Hence, Probabilistic Classifier Chains (PCC) were proposed which replaced the greedy strategy with exhaustive search \cite{Dembczynski2010PCC}, $\epsilon$-approximate search \cite{dembczynski2012label}, beam search \cite{kumar2013beam} or A* search \cite{mena2015using}. To address the issue of error propagating in CC, Ensemble of Classifier Chains (ECC) \cite{liu2015optimality} averages several predictions by different chains to improve the prediction. 



\textbf{Conditional Random Field} (CRF) \cite{Lafferty2001CRF} offers a general framework for structured prediction based on undirected graphical models. When used in multi-label classification, CRF suffers from the problem of computational intractability. To remedy this issue, \cite{ZhouZhihua2011MultiLabelCRF} applied ensemble methods and \cite{deng2014large} proposed a special CRF for problems involving specific hierarchical relations. In addition, \cite{Guo2011MultiCDN} proposed using Conditional Dependency Networks, although their method also depended on the Gibbs sampling for approximate inference.

\textbf{Ecological Models} Species distribution modeling has been studied extensively in ecology and \cite{elith2009species} presented a nice survey.
For single species models, \cite{Phillips2004Maxent} proposed max-entropy methods to deal with presence-only data. By taking imperfect detection into account, \cite{MacKenzie2004cooccurrence} proposed occupancy models, which were further improved with a stronger version of statistical inference \cite{Hutchinson2011}. Species distribution models have been extended to capture population dynamics using cascade models \cite{sheldon2011collective} and non-stationary predictor response models \cite{fink2013adaptive}.
%

Multi-species interaction models were also proposed \cite{Yu2011multispecies,harris2015generating}. Recently, Deep Multi-Species Embedding (DMSE) \cite{chen2016deep} uses a probit model coupled with a deep neural net to capture inter-species correlations. This approach is closely related to CRF and also requires MCMC sampling during training. 



\section{Experiments}

\subsection{Datasets and Implementation Details}
\begin{figure}[tb]
\centering
\includegraphics[width=0.75\linewidth]{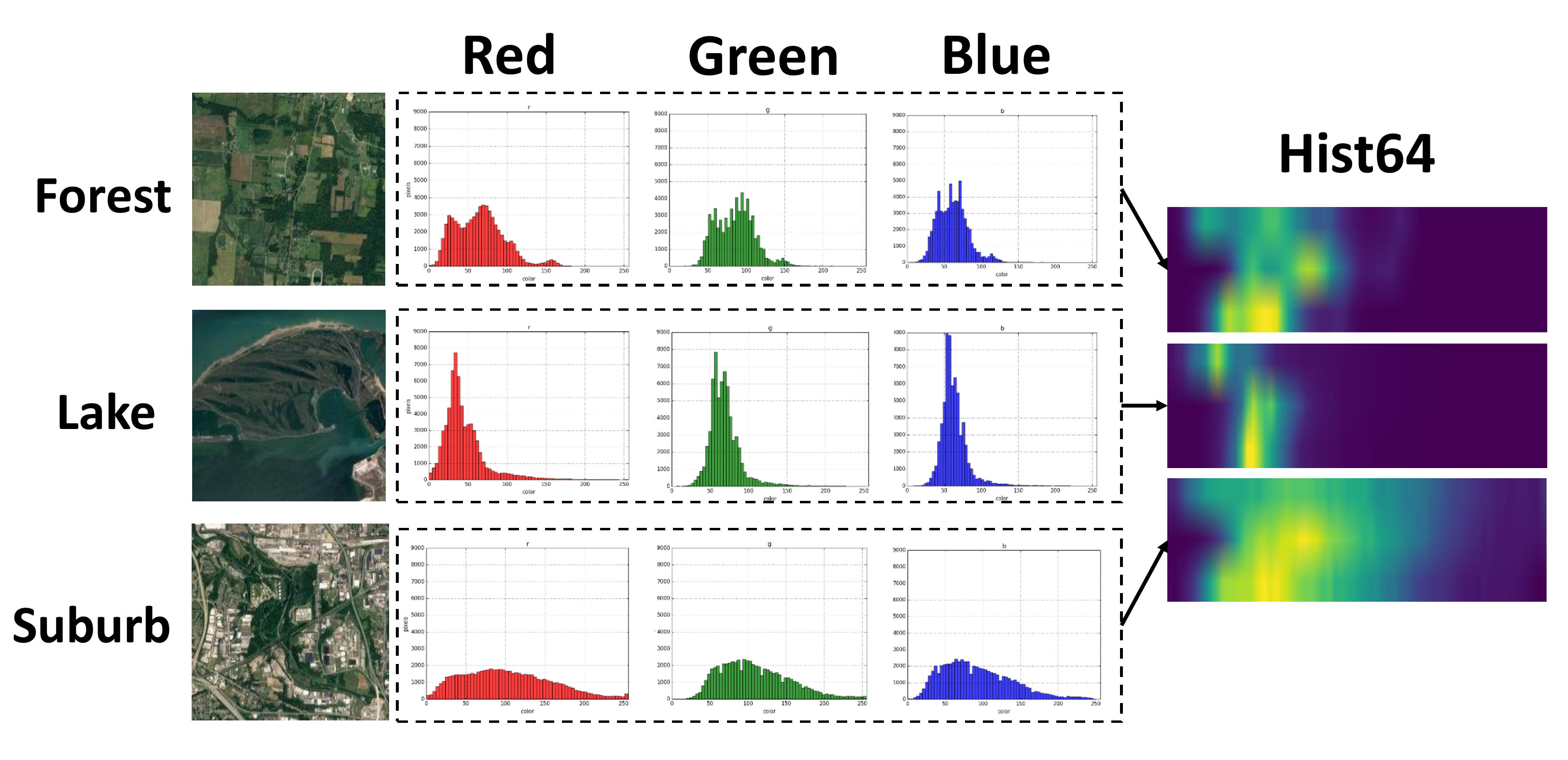}
\includegraphics[width=0.75\linewidth]{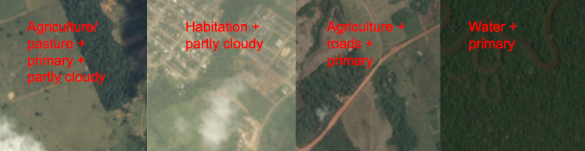}
\caption{(Top) Three satellite images (left) of different landscapes contain rich geographical information. We can also see that the histograms of RGB channels for each image (middle) contain useful information and are good summary statistics. (Bottom) Examples of sample satellite image chips and their corresponding landscape composition.}
\label{fig:hist}
\end{figure}

We evaluate our method on two computational sustainability related datasets. The first one is a crowd-sourced bird observation dataset collected from the \textit{eBird}  citizen science project \cite{munson2012ebird}. Each record in this dataset is referred to as a checklist in which the bird observer reports all the species he detects together with the time and the geographical location of an observational session. Crossed with the National Land Cover Dataset (NLCD) \cite{homer2015completion}, we get a 15-dimension feature vector for each location which describes the nearby landscape composition with 15 different land types such as water, forest, etc. In addition, to take advantages of rich external context information, we also collect satellite images for each observation site by matching the longitude and latitude of the observational site to Google Earth\footnote{https://www.google.com/earth/}. From the upper part of Figure \ref{fig:hist}, the satellite images of different geographical locations are quite different. Therefore these images contain rich geological information. Each image covers an area of 12.3$\mbox{km}^2$ near the observation site. For the use of training and testing, we transform all this data into the form ($x_i,y_i$), where $x_i$ denotes the contextual features including NLCD and satellite images and $y_i$ denotes the multi-species distribution. The whole dataset contains all the checklists from the Bird Conservation Region (BCR) 13 \cite{us2000north}  in the last two weeks of May from 2004 to 2014, which has 50,949 observations in total. Since May is a migration season and lots of non-native birds fly over BCR 13, this dataset provides a good opportunity to study these migratory birds using this dataset. We choose the top 100 most frequently observed birds as the target species which cover over 95\% of the records in our dataset. A simple mutual information analysis reveals rich correlation structure among these species.

Our second application is the Amazon rainforest landscape analysis\footnote{https://www.kaggle.com/c/planet-understanding-the-amazon-from-space} derived from Planet's full-frame analytic scene products. Each sample in this dataset contains a satellite image chip covering a ground area of 0.9 $\mbox{km}^2$. The chips were analyzed using the Crowd Flower\footnote{https://www.crowdflower.com/} platform to obtain ground-truth composition of the landscape. There are 17 composition label entities and they represent a reasonable subset of phenomena of interest in the Amazon basin and can broadly be broken into three groups: atmospheric conditions, common land cover phenomena and rare land use phenomena. Each chip has one or more atmospheric label entities and zero or more common and rare label entities.  
Sample chips and their composition are demonstrated in the lower part of Figure \ref{fig:hist}. 
There exists rich correlation between label entities, for instance, agriculture has a high probability to co-occur with water and cultivation. We randomly choose 34,431 samples for training, validation and testing. The details of the two datasets are listed in table \ref{table:dataset}. 

We propose two different neural network architectures for the \textit{feature network} to extract useful features from satellite images: multi-layer fully connected neural network (MLP) and convolutional neural network (CNN).
We also rescale images into different resolutions: Image64 for 64*64 pixels and Image256 for 256*256 pixels. In addition, we experiment using summary statistics such as the histograms of image's RGB channels (upper part of Figure \ref{fig:hist}) to describe an image (denoted as Hist). 
Inspired by \cite{you2017deep} that assumes permutation invariance holds and only the number of different pixel type in an image (pixel counts) are informative, we transfer each image into a matrix $\mathbf{H}\in\mathbb{R}^{d\times b}$, where $d$ is the number of band and $b$ is the number of discrete range section, thus $H_{i,j}$ indicates the percentage of pixels in the range of section $j$ for band $i$. We use RGB so $d=3$. We utilize histogram models with two different $b$ settings, Hist64 for $b=64$ and Hist128 for $b=128$. 

All the training and testing process of our proposed $\mathtt{MEDL\_CVAE}$ are performed on one NVIDIA Quadro 4000 GPU with 8GB memory. The whole training process lasts 300 epochs, using batch size of 512, Adam optimizer \cite{kingma2014adam} with learning rate of $10^{-4}$ and utilizing batch normalization \cite{ioffe2015batch}, 0.8 dropout rate \cite{srivastava2014dropout} and early stopping to accelerate the training process and to prevent overfitting.

\begin{table}[]
\newcommand{\tabincell}[2]{\begin{tabular}{@{}#1@{}}#2\end{tabular}}
\centering

\begin{tabular}{l|c|c|c}
\hline
\textbf{Dataset} & \textbf{Training Set Size} & \textbf{Test Set Size} & \textbf{\# Entities} \\ \hline
\textit{eBird} & 45855 & 5094 & 100 \\
\textit{Amazon} & 30383 & 4048 & 17 \\
\hline
\end{tabular}
\caption{the statics of the \textit{eBird} and the \textit{Amazon} dataset}
\label{table:dataset}
\end{table}

\subsection{Experimental Results}

\begin{table}[htb!]
\newcommand{\tabincell}[2]{\begin{tabular}{@{}#1@{}}#2\end{tabular}}
\centering
\begin{tabular}{l|c|c}
\hline
 \textbf{Method} & \textbf{Neg. JLL} & \tabincell{c}{\textbf{Time} \\ \textbf{(min)}} \\ \hline
NLCD+MLP & 36.32 & 2 \\
Image256+ResNet50 & 34.16 & 5.3 hrs \\ 
NLCD+Image256+ResNet50 & 34.48 & 5.7 hrs\\
NLCD+Hist64+MLP & 34.97 & 3\\
NLCD+Hist128+MLP & 34.62 & 4\\ 
NLCD+Image64+MLP & 33.73 & 9\\ \hline
NLCD+MLP+PCC & 35.99 & 21 \\
NLCD+Hist128+MLP+PCC & 35.07 & 33\\
NLCD+Image64+MLP+PCC & 34.48 & 53\\
NLCD+DMSE & 30.53 & 20 hrs\\ \hline
NLCD+MLP+$\mathtt{MEDL\_CVAE}$ & 30.86 & 9 \\
NLCD+Hist64+MLP+$\mathtt{MEDL\_CVAE}$  & 28.86 & 20\\
NLCD+Hist128+MLP+$\mathtt{MEDL\_CVAE}$ & 28.71 & 22\\
NLCD+Image64+MLP+$\mathtt{MEDL\_CVAE}$ & \textbf{28.24} & 48\\
\hline
\end{tabular}
\caption{Negative joint log-likelihood and training time of models assuming independence (first section), previous multi-entity dependence models (second section) and our $\mathtt{MEDL\_CVAE}$ on the $\textit{eBird}$ test set. $\mathtt{MEDL\_CVAE}$ achieves lower negative log-likelihood compared to other methods with the same feature network structure and context input while taking much less training time. Our model is also the only one among joint models (second and third section) which achieves the best log-likelihood taking images as inputs, while other models must rely on summary statistics to get good but suboptimal results within a reasonable time limit.}
\label{table:results_ebird}
\end{table}

We compare the proposed $\mathtt{MEDL\_CVAE}$ with two different groups of baseline models. 
The first group is \textit{models assuming independence structures among entities}; i.e., the distribution of all entities are independent of each other conditioned on the feature vector.
Within this group, we have tried models with different feature inputs, including models with highly advanced deep neural net structure, 
ResNet \cite{he2016identity}.  
The second group is \textit{previously proposed multi-entity dependence models} which have the ability to capture correlations among entities. Within this group, we compare with the recent proposed Deep Multi-Species Embedding (DMSE) \cite{chen2016deep}. This model is closely related to CRF, representing a wide class of energy based approach. Moreover, it further improves classic energy models, taking advantages of the flexibility of deep neural nets to obtain useful feature description. Nevertheless, its training process uses classic MCMC sampling approaches, therefore cannot be fully integrated on GPUs. 
We also compare with Probabilistic Classifier Chains (PCC) \cite{Dembczynski2010PCC}, which is a representative approach among a series of models proposed in multi-label classification. 

Our baselines and $\mathtt{MEDL\_CVAE}$ are all trained using different feature network architecture as well as satellite imagery with different resolution and encoding. We use Negative Joint Distribution Loglikelihood (Neg. JLL) as the main indicator of a model's performance: $-\frac{1}{N}\sum\limits_{i=1}^{N}\log Pr(y_i|x_i)$, where $N$ is the number of samples in the test set. For $\mathtt{MEDL\_CVAE}$ models, $Pr(y_i|x_i)=\mathtt{E}_{z_i\sim Pr(z_i | x_i)}\left[ Pr(y_i| z_i, x_i)\right]$. We obtain 10,000 samples from the posterior $Pr(z_i | x_i)$ to estimate the expectation. We also double checked that the estimation is close and within the bound of the variational lower bound in Equation~\ref{eq:vari}. 
The sampling process can be performed on GPUs within a couple of minutes. The experiment results on $\textit{eBird}$ and $\textit{Amazon}$ dataset are shown in Table \ref{table:results_ebird} and \ref{table:results_amazon}, respectively.

\begin{table}[htb!]
\newcommand{\tabincell}[2]{\begin{tabular}{@{}#1@{}}#2\end{tabular}}
\centering
\begin{tabular}{l|c}
\hline
 \textbf{Method} & \textbf{Neg. JLL} \\ \hline
Image64+MLP & 2.83  \\
Hist128+MLP & 2.44  \\
Image64+CNN & 2.16   \\ \hline
Image64+MLP+PCC & 2.95  \\
Hist128+MLP+PCC & 2.60  \\
Image64+CNN+PCC & 2.45  \\ \hline
Image64+MLP+$\mathtt{MEDL\_CVAE}$ & 2.37 \\
Hist128+MLP+$\mathtt{MEDL\_CVAE}$ & 2.09 \\
Image64+CNN+$\mathtt{MEDL\_CVAE}$ & \textbf{2.03} \\
\hline
\end{tabular}
\caption{Performance of baseline models and $\mathtt{MEDL\_CVAE}$ on the  \textit{Amazon} dataset. Our method clearly outperforms models assuming independence (first section) and previous multi-entity dependence models (second section) with various types of context input and feature network structures.}
\label{table:results_amazon}
\end{table}

We can observe that: (1){\bf MEDL\_CVAE significantly outperforms all independent models} given the same feature network (CNN or MLP) and context information (Image or Hist), even if we use highly advanced deep neural net structures such as ResNet in independent models. It proves that our method is able to capture rich dependency structures among entities, therefore outperforming approaches that assume independence among entities. (2) Compared with previous multi-entity dependence models, {\bf MEDL\_CVAE trains in an order of magnitude less time}. Using low-dimensional context information NLCD, which is a 15-dimensional vector, PCC's training needs nearly twice the time of $\mathtt{MEDL\_CVAE}$ and DMSE needs over 130 times (20 hours). (3) In each model group in Table~\ref{table:results_ebird}, it is clear that adding satellite images improves the  performance, which proves that {\bf rich context is informative.} (4) Only our model $\mathtt{MEDL\_CVAE}$  is able to take full advantage of the rich context in satellite images. Other models, such as DMSE, already suffer from a long training time with low-dimensional feature inputs such as NLCD, and cannot scale to using satellite images. 
It should be noted that NLCD+Image64+MLP+$\mathtt{MEDL\_CVAE}$ can achieve much  better performance with only 1/25 time of DMSE. PCC needs less training time than DMSE but doesn't perform well on joint likelihood. It is clear that {\bf due to the end-to-end training process on GPUs, our method is able to take advantage of rich context input to further improve multi-entity dependence modeling}, which is beyond the capacity of previous models.

\begin{figure}[tb]
\centering
\includegraphics[width=0.7\linewidth]{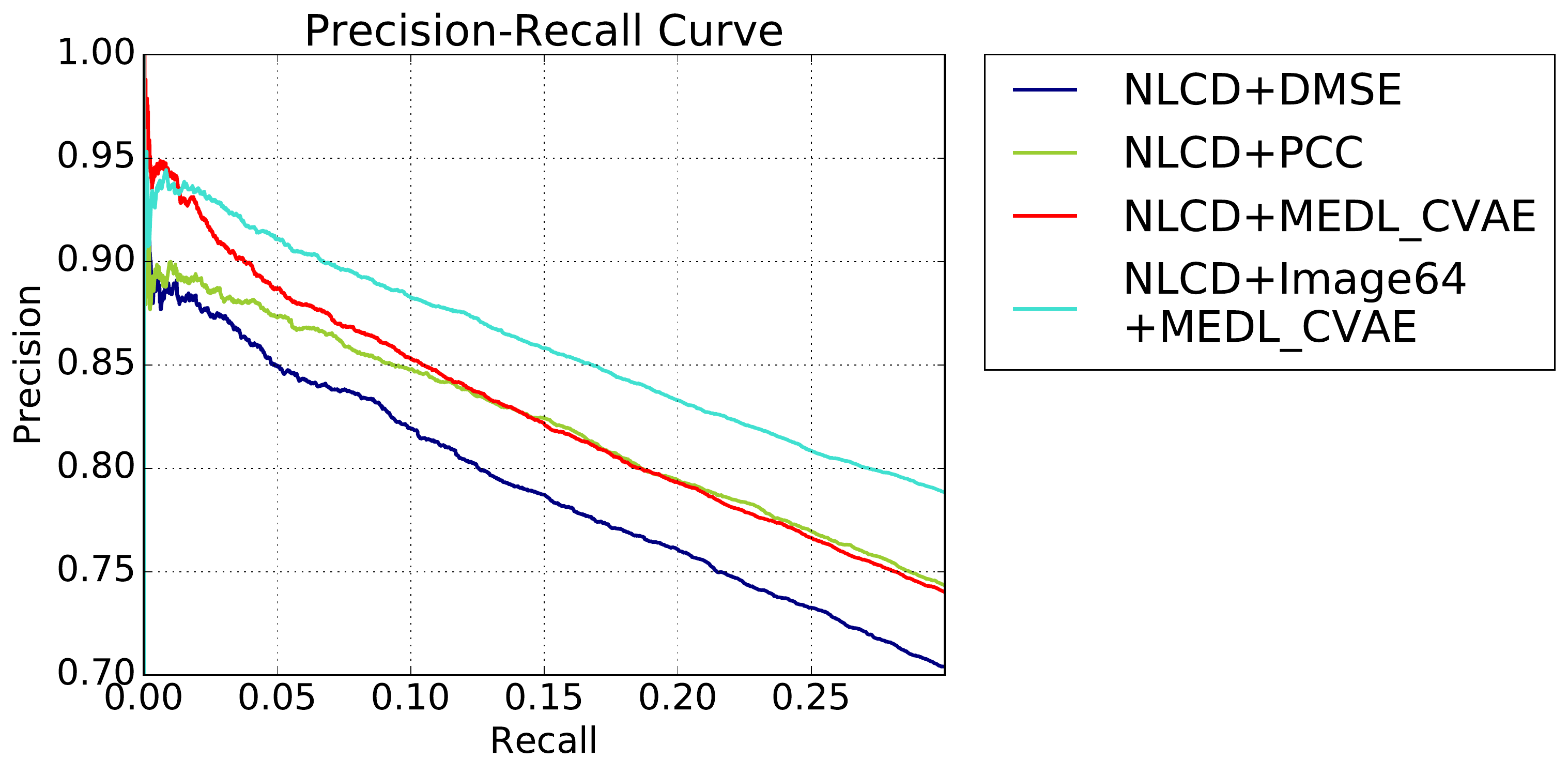}
\caption{Precison-Recall Curves for a few multi-entity dependence models on the $\textit{ebird}$ dataset (better view in color). $\mathtt{MEDL\_CVAE}$ utilizing images as input outperforms other joint methods without imagery.}
\label{fig:ebird_ap}
\end{figure}

To further prove $\mathtt{MEDL\_CVAE}$'s modeling power, we plot the precision-recall curve shown in Figure \ref{fig:ebird_ap} for all dependency models on the $\textit{ebird}$ dataset. The precision and recall is defined on the marginals to predict the occurrence of individual species and averaged among all 100 species in the dataset. As we can see, our method outperforms other models  after taking rich context information into account.

\begin{figure}[t]
    \begin{minipage}[t]{0.55\linewidth}
        \includegraphics[width=1.0\linewidth]{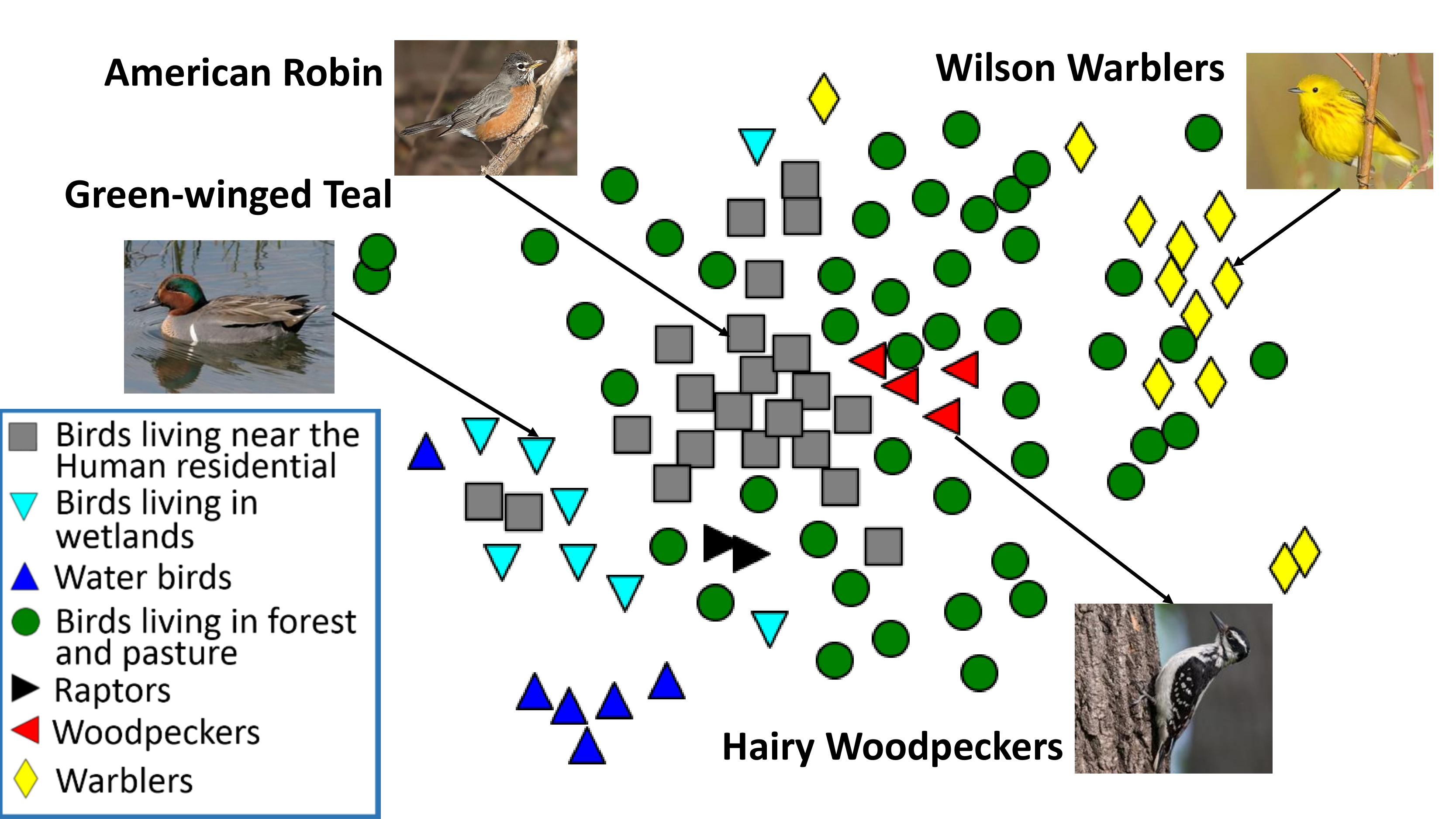}
        \captionsetup{labelformat=empty}
        \caption{(a) bird Embedding}
    \end{minipage}%
    \begin{minipage}[t]{0.44\linewidth}
        \centering
        \includegraphics[width=1.0\linewidth]{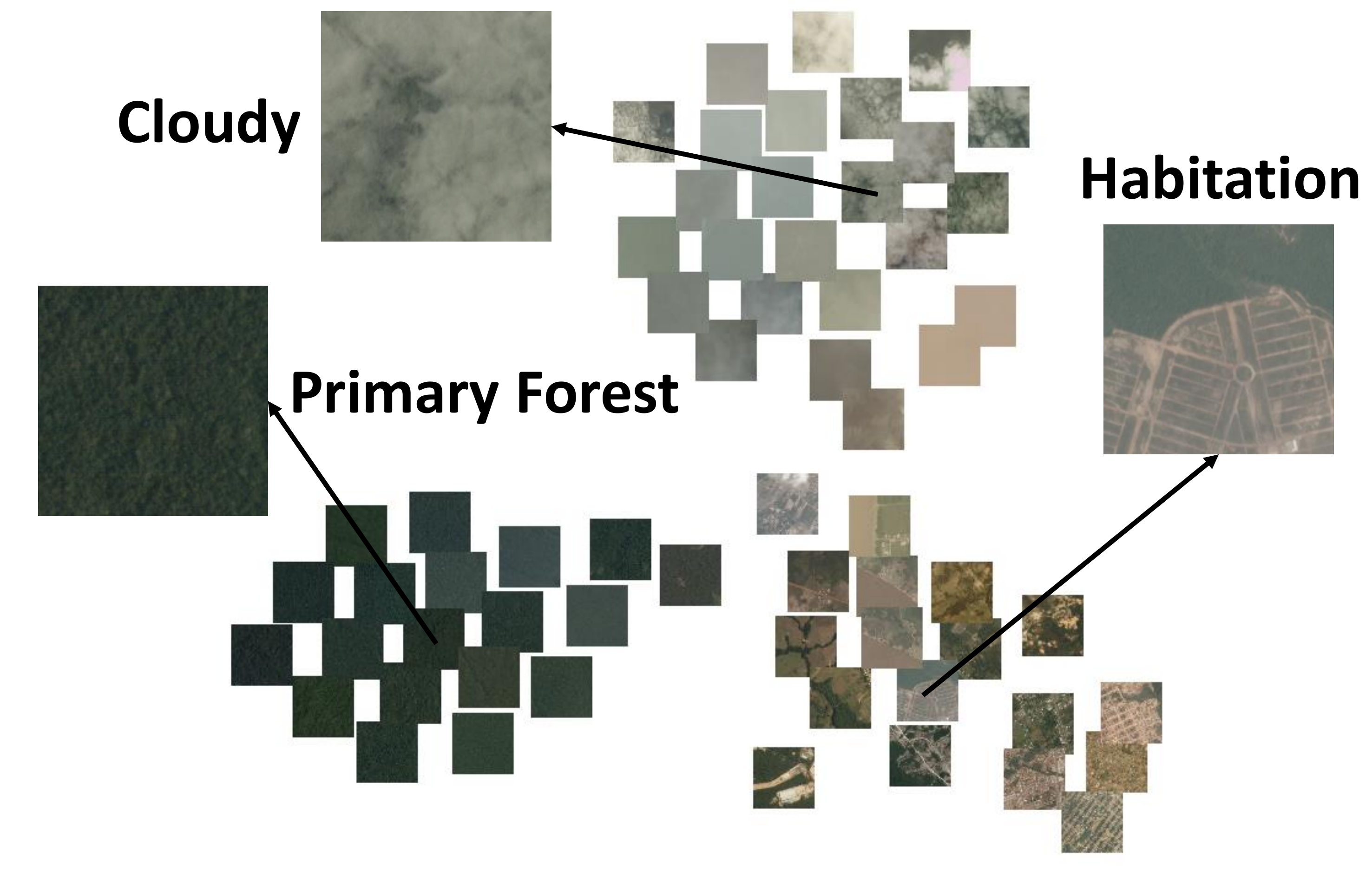}
        \captionsetup{labelformat=empty}
        \caption{(b) satellite image embedding}
    \end{minipage}
    \caption{(a) Visualization of the vectors inside decoder network's last fully connected layer gives a reasonable embedding of multiple bird species when our model is trained on the \textit{eBird} dataset. Birds living in the same habitats are clustered together. (b) Visualization of the posterior $z\sim Q(z|x,y)$ gives a good embedding on landscape composition when our model is trained on the \textit{Amazon} dataset.  Pictures with similar landscapes are clustered together.}
    \label{fig:embed}
\end{figure}

%

\subsubsection{Latent Space and Hidden Variables Analysis}

In order to qualitatively confirm that our $\mathtt{MEDL\_CVAE}$ learns useful dependence structure between entities, we analyze the latent space formed by the hidden variables in the neural network.
Inspired by \cite{chen2016deep}, each vector of decoder network's last fully connected layer can be treated as an embedding showing the relationship among species. Figure \ref{fig:embed} visualizes the embedding using t-SNE \cite{maaten2008visualizing}. We can observe that birds of the same category or having similar environmental preferences cluster together.
In addition, previous work \cite{kingma2013auto} has shown that the $\textit{recognition network}$ in Variational Auto-encoder is able to cluster high-dimensional data. Therefore we conjecture that the posterior of $z$ from the $\textit{recognition network}$ should carry meaningful information on the cluster groups. Figure \ref{fig:embed} visualizes the posterior of $z\sim Q(z|x,y)$ in 2D space using t-SNE on the $\textit{Amazon}$ dataset. We can see that satellite images of similar landscape composition also cluster together.

\section{Conclusion}

In this paper, we propose $\mathtt{MEDL\_CVAE}$ for multi-entity dependence learning, which encodes a conditional multivariate distribution as a generating process. As a result, the variational lower bound of the joint likelihood can be optimized via a conditional variational auto-encoder and trained end-to-end on GPUs. Tested on two real-world applications in computational sustainability, we show that $\mathtt{MEDL\_CVAE}$ captures rich dependency structures, scales better than previous methods, and further improves the joint likelihood taking advantage of very rich context information that is beyond the capacity of previous methods.
Future directions include exploring the connection between the current formulation of $\mathtt{MEDL\_CVAE}$ based on deep neural nets and the classic multivariate response models in statistics.

\bibliographystyle{plain} 
{\small
\bibliography{luming,multi_label,dichen2}
}

\end{document}